# Impact of Recurrent Neural Networks and Deep Learning Frameworks on Real-time Lightweight Time Series Anomaly Detection

Ming-Chang Lee[1], Jia-Chun Lin[2] and Sokratis Katsikas[3]

[1,2,3]Department of Information Security and Communication Technology, Norwegian University of Science and Technology,
Gjøvik, Norway

[1] mingchang1109@gmail.com
[2] jia-chun.lin@ntnu.no
[3] sokratis.katsikas@ntnu.no

# Impact of Recurrent Neural Networks and Deep Learning Frameworks on Real-time Lightweight Time Series Anomaly Detection

Ming-Chang Lee[0000−0003−2484−4366], Jia-Chun Lin[0000−0003−3374−8536], and Sokratis Katsikas[0000−0003−2966−9683]

Department of Information Security and Communication Technology, Norwegian University of Science and Technology (NTNU), Gjøvik, Norway
mingchang1109@gmail.com,{jia-chun.lin,sokratis.katsikas}@ntnu.no

**Abstract.** Real-time lightweight time series anomaly detection has become increasingly crucial in cybersecurity and many other domains. Its ability to adapt to unforeseen pattern changes and swiftly identify anomalies enables prompt responses and critical decision-making. While several such anomaly detection approaches have been introduced in recent years, they primarily utilize a single type of recurrent neural networks (RNNs) and have been implemented in only one deep learning framework. It is unclear how the use of different types of RNNs available in various deep learning frameworks affects the performance of these anomaly detection approaches due to the absence of comprehensive evaluations. Arbitrarily choosing a RNN variant and a deep learning framework to implement an anomaly detection approach may not reflect its true performance and could potentially mislead users into favoring one approach over another. In this paper, we aim to study the influence of various types of RNNs available in popular deep learning frameworks on real-time lightweight time series anomaly detection. We reviewed several state-of-the-art approaches and implemented a representative anomaly detection approach using well-known RNN variants supported by three widely recognized deep learning frameworks. A comprehensive evaluation is then conducted to analyze the performance of each implementation across real-world, open-source time series datasets. The evaluation results provide valuable guidance for selecting the appropriate RNN variant and deep learning framework for real-time, lightweight time series anomaly detection.

## 1 Introduction

A time series is known as a sequence of data points or observations taken or recorded through repeated measurements over time [2]. These observations can encompass a wide range of variables, including network traffic volume, system resource usage, retail sales, electricity consumption, weather conditions including temperature and humidity, and environmental factors like CO2 levels.

With the growing prevalence of the Internet of Things (IoT), a multitude of time series data is continuously generated by diverse IoT sensors and devices. Analyzing this time series data and detecting anomalies is of great importance to businesses and organizations, as it helps not only in identifying patterns and trends but also in detecting potential anomalies and security threats. This enables businesses and organizations to implement effective policies and security measures, thereby enhancing decision-making processes [18,37].

Time series anomaly detection aims to pinpoint and identify data points that deviate from the expected pattern or normal behavior within a time series, and it has been widely applied in various domains, such as cybersecurity [3,5], cloud infrastructure [11], smart grid operation [39], healthcare systems [31], and agricultural practices [28]. It is essential and desirable that time series anomaly detection is capable of accurately detecting anomalies in real time, conducting anomaly detection in a lightweight manner, and adapting to minor pattern changes without any offline model training, supervised learning, extensive human intervention or domain knowledge [21,24].

Many approaches for detecting anomalies in time series have been introduced in the past decade. Some are tailored for univariate time series, which involve only one time-dependent variable, while others are designed for multivariate time series, which consisting of multiple time-dependent variables. In this paper, our research focuses on univariate time series anomaly detection, serving as the fundamental building block for multivariate time series analysis [22]. To be more precise, our focus lies in univariate time series anomaly detection approaches that exhibit the following desired characteristics: online model training, unsupervised learning, real-time detection, lightweight design, adaptability, and minimal reliance on human intervention or domain knowledge [21]. These characteristics are imperative in determining the practicality and effectiveness of any approach in the context of time series anomaly detection [6].

According to our investigation, only a few state-of-the-art approaches satisfy the aforementioned characteristics. However, these approaches are often implemented using a single type of RNN, such as Long Short-Term Memory (LSTM), and typically within a specific deep learning (DL) framework. In reality, several DL frameworks have been introduced and are widely used, including TensorFlow [1], PyTorch [30], and Deeplearning4j [10]. They have a common goal to facilitate complicated data analysis process and offer integrated environments on top of standard programming languages [29]. Although a number of surveys and analyses have been conducted to compare different DL frameworks, they have primarily focused on either specific tasks (e.g., natural language processing) or different types of computing environments. A closely related study to our work was conducted by Lee and Lin [20]. In their work, the authors found that DL frameworks significantly impact real-time lightweight time series anomaly detec-tion approaches in terms of detection accuracy and time consumption. However, their study did not take the impact of different RNN variants into consideration.

To provide a comprehensive evaluation of how different RNN variants and DL frameworks impact real-time lightweight time series anomaly detection, this paper studied several state-of-the-art approaches with these characteristics. We then implemented the most representative approach using different RNN variants across three different DL frameworks. A series of experiments based on open-source, real-world time series data were performed to evaluate all the implementations. The results demonstrate that the choice of RNN variants and DL frameworks significantly influences both anomaly detection accuracy and time efficiency. Therefore, careful consideration of the selection of RNN variants and DL frameworks is crucial when designing and implementing adaptive, real-time, and lightweight time series anomaly detection approaches.

The rest of this paper is structured as follows: Section 2 introduces various RNN variants and DL frameworks. Section 3 provides an overview of the related work. Section 4 presents state-of-the-art real-time lightweight anomaly detection approaches and introduce the approach selected for our evaluation. Section 5 details our evaluation setup, followed by the evaluation results presented in Section 6. Finally, Section 7 concludes this paper and outlines future work.

## 2 RNN variants and DL frameworks

In this section, we introduce several RNN variants and well-known DL Frameworks.

### 2.1 RNN variants

A RNN [15] is a type of artificial neural network designed for processing sequential data or time series. Unlike traditional feedforward neural networks, RNNs have connections that loop back on themselves, allowing them to maintain a hidden state or memory of previous inputs. This recurrent structure makes RNNs well-suited for tasks involving sequential or time series data. In an RNN, a time step is processed one at a time, meaning that the network handles each data point sequentially and updates its internal state based on the current input and the previous state. This enables RNNs to capture dependencies and patterns across different time steps. However, RNNs have difficulties capturing long-term dependencies and might suffer from the vanishing gradient problem, which hinders their ability to learn from distant past inputs [14].

LSTM [13] is a type of RNN that was specifically designed to capture long-term dependencies and model temporal sequences. The structural framework of an LSTM closely resembles that of conventional RNN, with a key distinction being the presence of memory blocks as nonlinear units within each hidden layer. Each memory block operates autonomously, containing its dedicated memory cells and is equipped with three gates: the input gate, the output gate, and the forget gate. The use of these gates enables LSTM to combat the vanishing gradient problem [14], as it allows gradients to flow unchanged.

Gated Recurrent Unit (GRU) is an RNN architecture proposed by Cho et al. [8] to enable recurrent units to adaptively capture dependencies at various time scales. Similarly to LSTM, GRU employs gates to control information flow within the memory unit. However, it lacks an output gate, resulting in fewer parameters than LSTM. Chung et al. [9] evaluated LSTM and GRU in the context of sequence modeling using various datasets, such as polyphonic music and raw speech signals. Despite their efforts, they were unable to draw a definitive conclusion regarding whether LSTM or GRU performs better.

### 2.2   DL frameworks

TensorFlow [1] is an open-source DL framework developed by the Google Brain team, and it is one of the most popular and widely used DL frameworks. TensorFlow employs dataflow graphs to encapsulate both the computational logic within an algorithm and the corresponding state upon which the algorithm operates, meaning that users can define the entire computation graph before executing it. TensorFlow supports a wide range of neural network architectures and can leverage hardware acceleration using graphics processing units (GPUs) to accelerate model training and inference for both small-scale and large-scale applications. However, it is important to note that TensorFlow's complexity stems from its low-level API, which poses challenges to its user-friendliness. To enhance its user-friendliness and accessibility for a broader range of users, TensorFlow is often used in conjunction with Keras [16], a popular Python wrapper library known for providing a high-level, modular, and user-friendly API.

PyTorch [30] is an open-source deep learning framework that provides a flexible and user-friendly environment for developing and training machine learning models, especially neural networks. It is widely used in various artificial intelligence and deep learning applications, including computer vision and natural language processing. PyTorch distinguishes itself by incorporating a high-performance C++ runtime, allowing developers to leverage it for deployment in production environments and effectively bypass Python-driven inference [17]. PyTorch is also known for its dynamic computational graph, enabling flexible model architecture design and easier debugging. PyTorch places a strong emphasis on tensor computation with robust GPU acceleration capabilities.

Deeplearning4j is an open-source distributed deep learning framework, introduced by Skymind in 2014 [10,36]. This framework is exclusively designed for the Java programming language and the Java Virtual Machine (JVM) environment, and it is designed to bring deep neural networks and machine learning capabilities to the JVM ecosystem. Deeplearning4j is known for its scalability and compatibility with popular programming languages, allowing Java and Scala developers to build and train deep learning models. However, compared with PyTorch, Deeplearning4j presents a steeper learning curve due to its lower-level APIs and the need for a good understanding of Java and deep learning concepts. Additionally, the pace of development, updates, and the introduction of new features in Deeplearning4j may not be as rapid as in some other deep learning frameworks.

## 3 Related work

Several efforts have been made to compare DL frameworks. For examples, Kovalev et al. [19] conducted an evaluation in which they assessed the training time, prediction time, and classification accuracy of a fully connected neural network using five different DL frameworks: Theano with Keras, Torch, Caffe, TensorFlow, and Deeplearning4j. Zhang et al. [40] introduced a benchmark, encompassing six DL frameworks, different mobile devices, and fifteen DL models for image classification, object detection, semantic segmentation, and text classification. Their analysis shows that no single DL framework exhibits superiority across all tested scenarios. Additionally, they highlighted that the influence of DL frameworks may surpass both DL algorithm design and hardware capacity considerations. Despite the valuable insights provided by their research, their findings are unable to address our specific question regarding the influence of different RNNs and DL frameworks on real-time lightweight time series anomaly detection.

Zhang et al. [41] performed a comprehensive performance assessment of several DL frameworks, including TensorFlow, TensorFlow Lite, PyTorch, Caffe2, and MXNet, across diverse hardware platforms. The authors selected two different scales of convolutional neural network (CNN) models, and compared the performance of these models across various combinations of hardware and DL frameworks, focusing on metrics such as latency, memory footprint, and energy consumption. Based on the evaluation results, there is not a definitive winner for every metric, as each framework excels in some metrics. In addition, Zahidi et al. [38] conducted an analysis aimed at comparing various DL frameworks based on Python and Java. Their study specifically focused on assessing how these libraries facilitate natural language processing (NLP) tasks. However, it is worth mentioning that the CNN models and NLP tasks used in the two aforementioned papers are considerably more complex than lightweight time series anomaly detection models. Therefore, their findings and recommendations may not be applicable to our study.

Nguyen et al. [29] conducted a survey on various DL frameworks, where they analyzed the strengths and weaknesses of each library. However, their endeavor did not include the execution of experimental comparisons among these DL frameworks. Another similar comparison was carried out by Wang et al. [36], in which several DL frameworks were assessed, including their interface properties, deployment capabilities, performance, framework designs, etc. While the authors provided recommendations on selecting DL frameworks for different scenarios, their evaluation do not directly address the specific question that this paper aims to answer, which concerns the impact of RNN variants and DL frameworks on real-time lightweight time series anomaly detection.

# 4  Time series anomaly detection approaches and the selected approach for evaluation

In this section, we introduce several state-of-the-art time series anomaly detection approaches that are real-time lightweight. We then describe the specific approach we have chosen for further evaluation.

## 4.1  Time series anomaly detection approaches

Anomaly detection for univariate time series can broadly be categorized into two main types: statistical-based approaches and machine learning-based approaches according to [21]. Statistical-based approaches aim to create a statistical model that represents normal time series data and utilizes this model to identify anomalous data points in the time series. Notable examples of such approaches include AnomalyDetectionTs and AnomalyDetectionVec, developed by Twitter [34], and Luminol introduced by LinkedIn [27]. However, statistical-based approaches may have limitations, especially when dealing with data that does not conform to a known distribution [4]. Therefore, these approaches do not meet the criteria of being adaptive, even though they are generally considered lightweight.

In contrast, machine learning-based approaches are designed to identify anom -alies without the need to assume a specific model since they do not require knowledge of the underlying process data generation process [7]. RePAD [24] is a real-time anomaly detection approach for univariate time series based on LSTM and the Look-Back and Predict-Forward strategy. RePAD does not require any offline model training. Instead, it trains a simple LSTM model using the most recent historical data points and then uses the model to predict the next data point. RePAD evaluates whether the current LSTM model should be re-trained based on the difference between the actual values and predicted values compared with a dynamically calculated detection threshold at every time point. This design not only enables RePAD to adapt to minor pattern changes but also to detect anomalous data points in real-time. Furthermore, the simplicity of the LSTM architecture makes RePAD a lightweight approach without consuming considerable resources.

ReRe [23] is an enhanced time series anomaly detection approach that builds upon RePAD. Its primary objective is to mitigate high false positives introduced by RePAD. ReRe incorporates a dual-LSTM model approach to jointly identify anomalous data points. Both model operates similarly to RePAD, but the second model adopts a stricter detection threshold. In contrast to RePAD, ReRe requires slightly more computational resources, primarily due to its utilization of two LSTM models. SALAD [26] stands as another online, adaptive, unsupervised time series anomaly detection approach, specifically designed for time series exhibiting recurrent data patterns. It shares its foundation with RePAD, yet SALAD employs a two-phase methodology. In the initial phase, SALAD transforms the target time series into a sequence of average absolute relative error (AARE) values in real-time. Subsequently, in the second phase, it predicts an AARE value based on the most recent historical AARE values. If the difference between a real AARE value and its corresponding forecast AARE value exceeds a self-adaptive detection threshold, the associated data point is considered anomalous. The evaluation results shows SALAD provides higher detection accuracy than RePAD and ReRe, especially when dealing with recurrent time series. However, due to the employment of the two phases, SALAD requires more computational resources and more processing time for detecting anomalies.

Lee and Lin [21] identified a potential resource exhaustion issue in RePAD when applied to open-ended time series, especially over extended periods. ReRe and SALAD based on RePAD might have the same issue for open-ended time series. In response to this issue, they introduced RePAD2, which addresses the issue by redesigning the self-adaptive detection threshold to better accommodate open-ended time series. Their evaluation results demonstrate that RePAD2 achieves comparable detection performance to RePAD, affirming that RePAD2 still possesses adaptive, real-time, and lightweight characteristics. However, the impact of various RNNs and DL frameworks on RePAD2 has not been investigated, as RePAD2 is solely implemented using LSTM in Deeplearning4j. The same situation also occurs in RePAD, ReRe, and SALAD, as all of them were implemented using LSTM in Deeplearning4j.

RoLA [22] represents an advanced real-time anomaly detection system designed, but it is designed for multivariate time series data. In RoLA, each univariate time series within a target multivariate time series is separately processed by an anomaly detector that is built upon RePAD2. When an anomaly detector detects a suspicious data point, RoLA employs a majority rule to collectively determine whether that data point is anomalous or not by considering the correlations of all variables within a recent time period. Similar to all the other abovementioned approaches, RoLA was implemented using LSTM in Deeplearning4j.

Based on the above discussion, in this paper, we chose to focus on RePAD2 as our study target for three primary reasons. First, RePAD2 is fundamentally identical to RePAD, which serves as a building block for many state-of-the-art adaptive, real-time, and lightweight time series anomaly detection approaches. Second, RePAD2 effectively addresses the resource exhaustion problem that RePAD encountered while preserving comparable detection performance. In the next subsection, we will introduce RePAD2 and provide a detailed description of its design.

## 4.2 RePAD2

RePAD2 is designed to detect anomalous data points from an open-ended time series in real time without any offline model training. Let $T$ denote the current time point, starting from 0, which indicates the first time point in the target time series. RePAD2 trains an LSTM model using three historical data points and then utilizes this model to predict the next upcoming data point. Due to this design, the first LSTM model can be trained at time point 2, and the second LSTM model can be trained at time point 3. In order to identify anomalies,

RePAD2 uses Equation 1 to calculate an AARE value at every $T$, denoted by $AARE_T$.

$$AARE_T = \frac{1}{3} \sum_{y=T-2}^{T} \frac{\mid D_y - \widehat{D_y} \mid}{D_y}, T \geq 5 \qquad (1)$$

$D_y$ and $\widehat{D_y}$ denote the actual and predicted data point values at time point $y$, respectively. A low AARE value indicates that the predicted values closely match the observed values. Furthermore, to calculate its detection threshold $thd$ (see Equation 2), RePAD2 requires a minimum of three AARE values, enabling the calculation of thresholds at each time point from time point 7 onward. In Equations 3 and 4, parameter $W$ is utilized to constrain the number of historical AARE values used in calculating $thd$. If the total number of historical AARE values is less than $W$, all values are used; otherwise, only the $W$ most recent ones are used to calculate $thd$, thereby preventing resource exhaustion.

$$thd = \mu_{aare} + 3 \cdot \sigma_{aare}, T \geq 7 \qquad (2)$$

$$\mu_{aare} = \begin{cases} \frac{1}{T-4} \sum_{x=5}^{T} AARE_x, 7 \leq T < W+4 \\ \frac{1}{W} \sum_{x=T-W+1}^{T} AARE_x, \ T \geq W+4 \end{cases} \qquad (3)$$

$$\sigma_{aare} = \begin{cases} \sqrt{\frac{\sum_{x=5}^{T}(AARE_x - \mu_{AARE})^2}{T-4}}, 7 \leq T < W+4 \\ \sqrt{\frac{\sum_{x=T-W+1}^{T}(AARE_x - \mu_{AARE})^2}{W}}, \ T \geq W+4 \end{cases} \qquad (4)$$

At every $T$ where $T \geq 7$, RePAD2 compares $AARE_T$ with the current $thd$. If $AARE_T$ does not surpass $thd$, the data point at $T$, denoted by $D_T$, is not considered anomalous, and the current LSTM model is preserved for future prediction. However, if $AARE_T \geq thd$, RePAD2 attempts to adapt to potential pattern changes by retraining another LSTM model with the three most recent data points to re-predict $D_T$. If the new model produces an AARE value lower than $thd$, RePAD2 does not regard $D_T$ anomalous. Otherwise, RePAD2 immediately reports $D_T$ as anomalous, facilitating corresponding actions or countermeasures.

## 5   Evaluation setup

In this section, we provide a detailed description of our evaluation process for the target anomaly detection approach, RePAD2. Recall that RePAD2 was originally implemented using LSTM in Deeplearning4j. To understand the impact of various RNNs and DL frameworks on the performance of RePAD2, in this paper, we implemented RePAD2 using three different types of RNNs, namely RNN, LSTM, and GRU, and three different DL frameworks, namely TensorFlow-Keras, PyTorch, and Deeplearning4j. Our selection of TensorFlow-Keras and PyTorch is based on their well-established popularity and widespread adoption within the field. These frameworks have gained significant recognition and community support, making them ideal choices for our research. Considering that both TensorFlow-Keras and PyTorch are Python-based, it would be interesting to investigate the impact of Deeplearning4j in comparison to TensorFlow-Keras and PyTorch.

In our evaluation, the versions of TensorFlow-Keras, PyTorch, and Deeplearning4j are 2.9.1, 1.13.1, and 0.7-SNAPSHOT, respectively. It is important to note that Deeplearning4j officially supports only the LSTM architecture; it does not support RNN or GRU. Consequently, we implemented RePAD2 using the LSTM architecture within the Deeplearning4j framework. We refer to this specific implementation as DL4J-LSTM, which denotes the use of LSTM in Deeplearning4j for RePAD2. On the other hand, PyTorch officially supports RNN, LSTM, and GRU. These implementations are referred to as PT-RNN, PT-LSTM, and PT-GRU in the paper, respectively. Similarly, TensorFlow-Keras supports RNN, LSTM, and GRU, and these implementations are denoted as TFK-RNN, TFK-LSTM, and TFK-GRU in the paper. In total, we provide seven implementations, which are listed in Table 1. The term in the paper 'N/A' indicates that an implementation is not available.

**Table 1.** The seven implementations studied in this paper.

|      | TensorFlow-Keras | PyTorch  | Deeplearning4j |
|------|------------------|----------|----------------|
| RNN  | TFK-RNN          | PT-RNN   | N/A            |
| LSTM | TFK-LSTM         | PT-LSTM  | DL4J-LSTM      |
| GRU  | TFK-GRU          | PT-GRU   | N/A            |

### 5.1 Real-world datasets

To evaluate the seven implementations, three real-world time series datasets related to air quality from the UC Irvine Machine Learning Repository [35] were used. The first time series is called 'PT08.S1(CO)', representing the hourly averaged sensor response, specifically targeting carbon monoxide. The second time series is 'C6H6(GT)', which denotes the true hourly averaged Benzene concentration in microg/$m^3$. The last time series, 'PT08.S2(NMHC)', represents the hourly averaged sensor response primarily focused on non-methane hydrocarbons. Each of these time series consists of 9,357 data points, including 3 individual missing data points and 13 instances of collective missing data points. In the original dataset, each missing point was represented by a value of $-200$. To enhance the readability of this paper, we renamed these three time series as PT08.S1, C6H6, and PT08.S2. Table 2 summarizes the details of these time series.

Given that missing data points may indicate sensor failures or malfunctions, in this paper, we consider each individual missing point as a point anomaly, and each instance of collective missing points as a collective anomaly. Note that a point anomaly is defined as a single data point that deviates from the rest of the time series, while a collective anomaly consists of a sequence of data points that together form an anomalous pattern [33]. We replayed each of the three abovementioned time series as a stream and injected them to each implementation to evaluate how well these implementations can detect anomalies without prior knowledge of the time series.

Table 2. Details of three real-world time series used in our evaluation.

| Name | Data points | Interval | Duration | Anomalies |
|---|---|---|---|---|
| PT08.S1 | 9,357 | 1 hr | 2004/03/10 18:00 to 2005/04/04 14:00 | 3 points anomalies and 13 collective anomalies |
| C6H6 | 9,357 | 1 hr | 2004/03/10 18:00 to 2005/04/04 14:00 | 3 point anomalies and 13 collective anomalies |
| PT08.S2 | 9,357 | 1 hr | 2004/03/10 18:00 to 2005/04/04 14:00 | 3 point anomalies and 13 collective anomalies |

### 5.2  Hyperparameters, Parameters, and Environment

To guarantee a fair and impartial evaluation, all the seven implementations were configured with identical hyperparameters and parameters, as detailed in Table 3. This aligns with the settings employed by RePAD [24] and RePAD2 [21]. As mentioned earlier, RePAD2 utilizes the Look-Back and Predict-Forward strategy to determine the data size for online model training and prediction. In our study, we configured the Look-Back parameter and the Predict-Forward parameter to values of 3 and 1, respectively. This choice aligns with the recommendations suggested by [25]. In other words, each implementation always uses the three most recent data points to train an LSTM model, which is then used to predict the next data point.

Furthermore, the seven implementations inherited the simple structure of the recurrent neural network used by RePAD2 [21], namely, only one hidden layer with ten hidden units. It is also important to note that early stopping [12], which can automatically determine the number of epochs to prevent the LSTM models from overfitting the data, was not officially supported by PyTorch at the time of conducting the evaluation. Therefore, none of the implementations employed early stopping; instead, their epoch parameters were uniformly set to 50 for fairness and consistency.

In addition, recall that RePAD2 employs the parameter $W$ to mitigate the potential issue of resource exhaustion over extended periods. According to the experiment results of RePAD2 [21], setting a large value for $W$ is recommended as it helps reduce false positives and increase F1-score of RePAD2. Given the limited size of each chosen time series in this paper, $W$ was consistently set to the length of the respective time series when evaluating each implementation.

Table 3. Hyperparameter and parameter settings used by each implementation.

| Hyperparameters/parameters | Value |
|---|---|
| The Look-Back parameter | 3 |
| The Predict-Forward parameter | 1 |
| The number of hidden layers | 1 |
| The number of hidden units | 10 |
| The number of epochs | 50 |
| Learning rate | 0.005 |
| Activation function | tanh |
| Random seed | 140 |

The evaluation of each implementation on the three aforementioned time series was separately executed on a MacBook running MacOS 14.4.1. This ma-chine is equipped with a 2.6 GHz 6-Core Intel Core i7 processor and 16GB DDR4 SDRAM. It is imperative to underscore that the decision to conduct the evaluation on such a commodity computer, without GPUs or high-performance computing resources, was deliberate. This decision aims to assess how the combination of RNN variants and DL frameworks impacts the performance of RePAD2 in a generic environment.

## 6 Evaluation results

To evaluate the detection accuracy of each implementation, we considered precision (defined as $\frac{TP}{TP+FP}$), recall (defined as $\frac{TP}{TP+FN}$), and F1-score (defined as $2 \cdot \frac{precision \cdot recall}{precision+recall}$) where $TP$, $FP$, and $FN$ represent true positives, false positives, and false negatives, respectively. Precision measures the accuracy of positive predictions made by a model, while recall (also known as sensitivity) measures the model's ability to correctly identify all positive instances. The F1-score (also known as F-score) summarizes a model's performance in terms of both making accurate positive predictions and capturing all actual positives. A higher F1-score indicates better detection accuracy.

In addition, we incorporated the evaluation method used by [23] to measure $TP$, $FP$, and $FN$. If any point anomaly occurring at time point $A$ can be detected within the time period from $A-K$ to $A+K$, this anomaly is considered correctly detected. On the other hand, for any collective anomaly, if it starts at time point $C$ and ends at time point $D$ ($D>C$), and it can be detected within the period from $C-K$ to $D$, then this anomaly is considered correctly detected. Note that we adhered to the approach described in [32] and set $K$ to 3 for hourly-interval time series. This setting was applied consistently to all the seven implementations.

Furthermore, we employed the following three performance metrics to evaluate the efficiency of each implementation.

- Online model retraining ratio: the proportion of data points requiring online model relative to the total data points in the time series. A lower ratio indicates more efficient resource utilization and generally faster detection, as model training requires time.
- Time taken to detect anomalies for each data point when model retraining is required (DT-Train). This includes the time for training a new prediction model as well as for prediction and anomaly detection.
- Time taken to detect anomalies for each data point when model retraining is not required (DT-noTrain). This also signifies that the detection process can take place immediately without any delay.

Table 4 lists the detection accuracy of each implementation across these three time series. When these implementations were individually applied to the PT08.S1 time series, only DL4J-LSTM, TFK-RNN, TFK-GRU, and PT-RNN successfully detected all anomalies, achieving a recall of 1 for each. However, as illustrated in Fig. 1, each implementation also generated a number of false positives. Among all implementations, DL4J-LSTM achieves the highest F1-score (0.991) because it not only detected all anomalies but also made fewer false positives than the other implementations.

**Table 4.** Detection accuracy of each implementation across three time series. Note that P, R, and F1 denotes precision, recall, and F1-score, respectively.

|          | PT08.S1 |       |       | C6H6  |       |       | PT08.S2 |       |       |
|----------|---------|-------|-------|-------|-------|-------|---------|-------|-------|
|          | P       | R     | F1    | P     | R     | F1    | P       | R     | F1    |
| DL4J-LSTM| 0.981   | 1     | **0.991** | 0.936 | 1     | **0.967** | 0.979   | 1     | **0.989** |
| TFK-RNN  | 0.951   | 1     | 0.975 | 0.893 | 1     | 0.943 | 0.948   | 1     | 0.973 |
| TFK-LSTM | 0.893   | 0.933 | 0.913 | 0.789 | 0.867 | 0.826 | 0.810   | 1     | 0.895 |
| TFK-GRU  | 0.941   | 1     | 0.970 | 0.843 | 0.633 | 0.886 | 0.861   | 1     | 0.925 |
| PT-RNN   | 0.931   | 1     | 0.964 | 0.876 | 0.933 | 0.903 | 0.876   | 0.933 | 0.903 |
| PT-LSTM  | 0.874   | 0.867 | 0.870 | 0.796 | 0.867 | 0.830 | 0.845   | 0.933 | 0.887 |
| PT-GRU   | 0.920   | 0.733 | 0.816 | 0.819 | 1     | 0.900 | 0.838   | 1     | 0.912 |

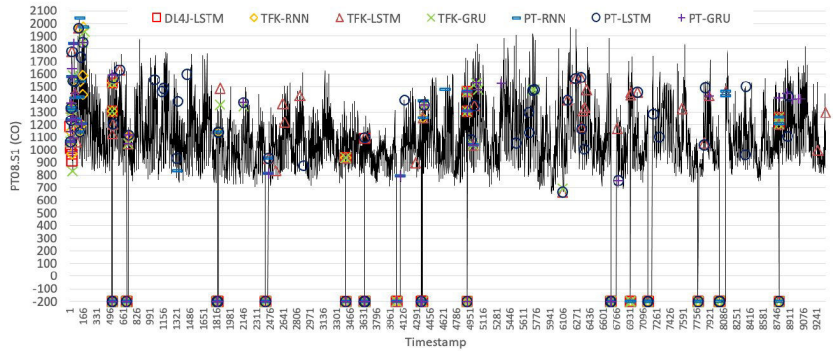

**Fig. 1.** Visualization of the PT08.S1 time series along with all data points detected as anomalous by each implementation.

To demonstrate this visually, we depicted all AARE values and detection threshold for each implementation on the PT08.S1 time series in Fig. 2, where each true anomaly is highlighted with a purple bar. It is evident that when DL4J-LSTM was tested, the AARE values at the time points of each true anomaly exceeded the threshold, allowing DL4J-LSTM to accurately identify these data points as anomalous, thereby achieving a recall of 1. Additionally, the AARE values at all the other time points are below the threshold, except for those occurring before the first anomaly. This explains why DL4J-LSTM reported fewer false positives compared to the other implementations and achieved a high precision of 0.981.

Furthermore, as shown in Table 4, TFK-RNN achieved the second highest F1-score (0.975) because it detected all anomalies but generated slightly more false positives than DL4J-LSTM. On the other hand, PT-GRU is the least effective among all implementations in detecting anomalies within the PT08.S1 time series, with the lowest recall of 0.733, leading to the lowest F1-score of 0.816. This low recall is easily observable in Fig. 2, which clearly shows that PT-GRU failed to identify more anomalies than the other implementations.

When the seven implementations were individually applied to the remaining two time series, namely the C6H6 and PT08.S2 time series, DL4J-LSTM con-sistently achieved the best detection accuracy due to its highest precision and recall (as shown Table 4). These results confirm that implementing RePAD2 using LSTM provided by Deeplearning4j offers the most effective and reliable anomaly detection. The second-best implementation is TFK-RNN. This implementation accurately detected all anomalies across all three time series, although it generated slightly more false positives than DL4J-LSTM. The remaining implementations exhibited unstable and varied detection accuracies across the three time series and introduced many false positives, as illustrated in Fig. 3 and Fig. 4. Consequently, they are not recommended for implementing RePAD2 or similar time series anomaly detection approaches.

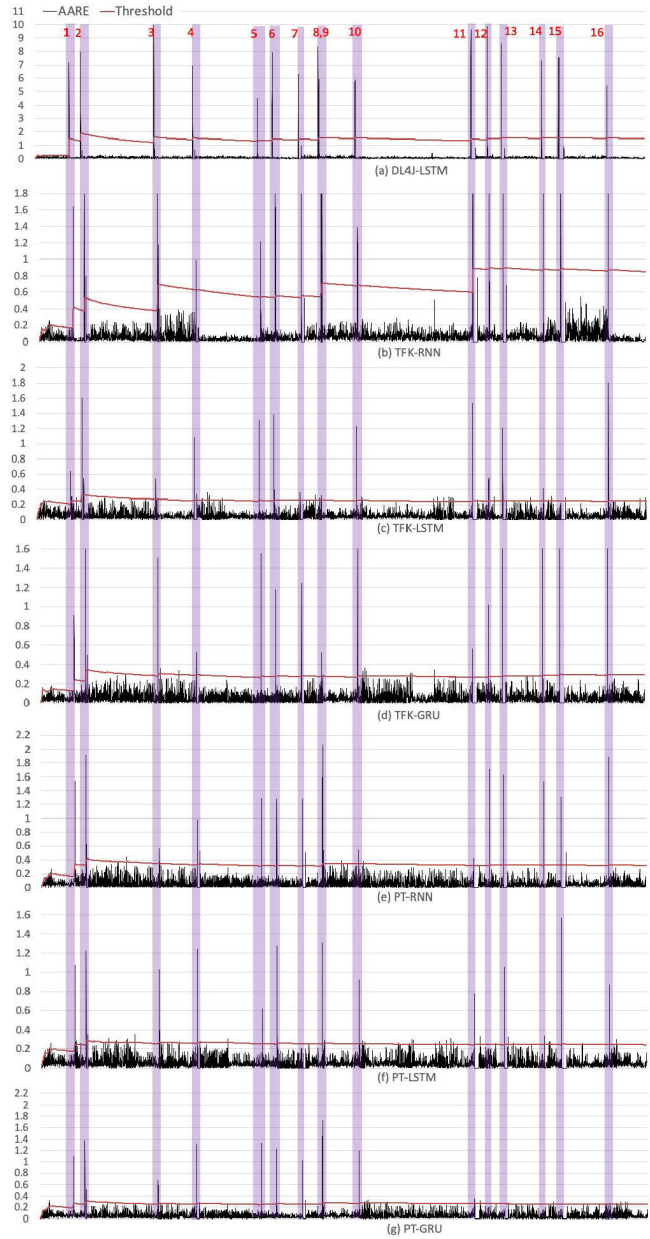

**Fig. 2.** AARE values and detection thresholds for each implementation on the PT08.S1 time series, with all true anomalies highlighted by a purple bar.

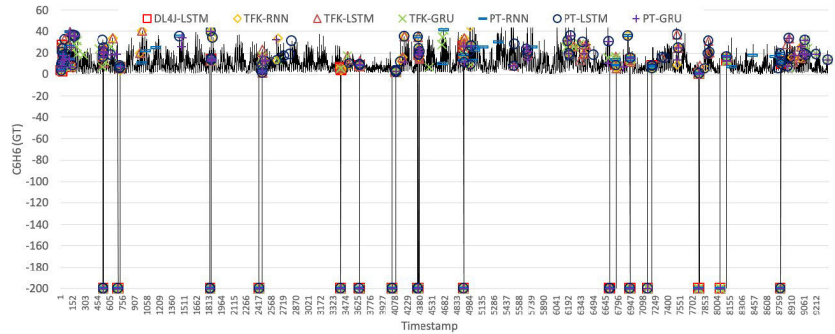

**Fig. 3.** Visualization of the C6H6 time series along with all data points detected as anomalous by each implementation.

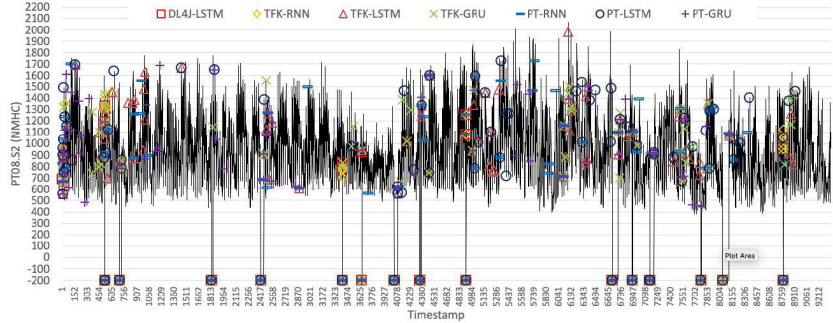

**Fig. 4.** Visualization of the PT08.S2 time series along with all data points detected as anomalous by each implementation.

Table 5 further displays the online model retraining ratios for the seven implementations across the three time series. Apparently, none of the implementations require significant model retraining. Among all implementations, both DL4J-LSTM and TFK-RNN achieved the lowest model training ratio across all three time series, indicating that their prediction models performed well. Consequently, there is no need to frequently replace them with new models. On the other hand, PT-LSTM and PT-GRU exhibited higher retraining ratios compared to the others, suggesting that their prediction models were less stable and thus required frequent model retraining and replacement.

Table 5. Online model retraining ratio of the seven implementations.

|  | PT08.S1 | C6H6 | PT08.S2 |
|---|---|---|---|
| DL4J-LSTM | **0.011** | **0.014** | **0.011** |
| TFK-RNN | **0.011** | **0.014** | **0.011** |
| TFK-LSTM | 0.017 | 0.021 | 0.026 |
| TFK-GRU | 0.013 | 0.021 | 0.021 |
| PT-RNN | 0.019 | 0.023 | 0.025 |
| PT-LSTM | 0.022 | 0.028 | 0.027 |
| PT-GRU | 0.018 | 0.028 | 0.029 |

Regarding time consumption in anomaly detection, as shown in Table 6 and Table 7, it is evident that every implementation exhibits a DT-Train longer than DT-noTrain across all time series. This is expected because DT-Train includes not only prediction and anomaly determination time but also online model retraining time. Conversely, DT-noTrain encompasses only prediction and anomaly determination time since model retraining is not required. Apparently, the three implementations based on PyTorch required significantly less time than those based on Deeplearning4j and TensorFlow-Keras when model retraining was required (see Table 6). This finding highlights the superior efficiency of the PyTorch framework in model retraining scenarios. Additionally, DL4J-LSTM offers the second-best time efficiency. However, all TensorFlow-Keras based implementations are more time-consuming.

Table 6. DT-Train of each implementation (sec).

|  | PT08.S1 | | C6H6 | | PT08.S2 | |
|---|---|---|---|---|---|---|
|  | Avg. | Std. | Avg. | Std. | Avg. | Std. |
| DL4J-LSTM | 0.267 | 0.029 | 0.255 | 0.012 | 0.257 | 0.019 |
| TFK-RNN | 0.773 | 0.141 | 0.747 | 0.100 | 0.757 | 0.119 |
| TFK-LSTM | 1.891 | 0.252 | 1.831 | 0.237 | 1.904 | 0.253 |
| TFK-GRU | 1.823 | 0.233 | 1.857 | 0.271 | 1.974 | 0.322 |
| PT-RNN | **0.059** | **0.010** | **0.059** | **0.009** | **0.060** | **0.008** |
| PT-LSTM | 0.067 | 0.010 | 0.067 | 0.009 | 0.071 | 0.014 |
| PT-GRU | 0.069 | 0.012 | 0.067 | 0.008 | 0.066 | 0.009 |

Table 7. DT-noTrain of each implementation (sec).

|  | PT08.S1 | | C6H6 | | PT08.S2 | |
|---|---|---|---|---|---|---|
|  | Avg. | Std. | Avg. | Std. | Avg. | Std. |
| DL4J-LSTM | **0.014** | **0.003** | **0.013** | **0.002** | **0.013** | **0.002** |
| TFK-RNN | 0.172 | 0.032 | 0.166 | 0.032 | 0.169 | 0.033 |
| TFK-LSTM | 0.451 | 0.066 | 0.446 | 0.064 | 0.451 | 0.069 |
| TFK-GRU | 0.407 | 0.062 | 0.423 | 0.076 | 0.444 | 0.083 |
| PT-RNN | 0.022 | 0.003 | 0.022 | 0.003 | 0.022 | 0.002 |
| PT-LSTM | 0.021 | 0.002 | 0.021 | 0.002 | 0.021 | 0.002 |
| PT-GRU | 0.021 | 0.002 | 0.021 | 0.002 | 0.021 | 0.003 |

On the other hand, when model retraining was not required, DL4J-LSTM proves to be the most efficient implementation, as shown in Table 7. It took only approximately 0.013 to 0.014 seconds on average to determine the anomaly status of each data point within these three time series. Given that most data points did not require model retraining, as indicated by the low retraining ratio shown in Table 5, DL4J-LSTM can detect anomalies instantly upon receiving new data points. Meanwhile, the three implementations based on PyTorch rank as the second most efficient, while all TensorFlow-Keras implementations remain more time-consuming. Therefore, we conclude that TensorFlow-Keras may not be the ideal framework for implementing real-time lightweight time series anomaly detection approaches.

Based on all the evaluation results above, we conclude that adopting LSTM provided by Deeplearning4j is the most suitable choice for implementing RePAD2 and similar real-time lightweight time series anomaly detection approaches. This combination not only achieves outstanding detection accuracy but also maintains satisfactory detection efficiency. Additionally, while the PyTorch-based implementations offer the best time efficiency among all implementations when model retraining is required, they were unable to consistently provide satisfactory detection accuracy across all three time series. Finally, it is evident that TensorFlow with Keras is a less suitable option for implementing RePAD2 or similar anomaly detection approaches, due to its lower time efficiency and inconsistent detection accuracy.

## 7   Conclusions and future work

In this paper, we have systematically investigated the impact of RNN variants and DL frameworks on real-time lightweight time series anomaly detection. We examined state-of-the-art approaches that meet these criteria and implemented the most representative one, RePAD2, using three different types of RNNs (namely RNN, LSTM, and GRU) across three well-known DL frameworks (Deeplearning4j, TensorFlow-Keras, and PyTorch). All different implementations were thoroughly evaluated through a series of experiments using six performance metrics across three open-source time series datasets in total.

The experiment results demonstrate that RNN variants and DL frameworks have a significant impact on RePAD2 in terms of both detection accuracy and detection time efficiency. Therefore, it is crucial to carefully consider the choice of RNN variants and DL frameworks when designing real-time lightweight time series anomaly detection approaches to fully ascertain their true performance.

According to our evaluation, all RNN variants supported by TensorFlow-Keras are not recommended, as they required more time for anomaly detection than those based on Deeplearning4j and PyTorch. In other words, TensorFlow-Keras resulted in the longest detection times. However, if TensorFlow-Keras must be used for specific reasons, its basic RNN variant is recommended over others, as it provides better time efficiency and higher detection accuracy compared to other RNN variants. Additionally, our evaluation results indicate that all PyTorch-based implementations offer shorter detection times than those based on TensorFlow-Keras, while providing comparable efficiency to Deeplearning4j-based implementations, thus enabling real-time processing and instant responses. However, similar to the TensorFlow-Keras-based implementations, all PyTorch-based implementations exhibited unstable detection accuracy across all tested time series. If there is a specific need for using PyTorch, its RNN variant is the most recommended due to its better detection accuracy compared to other PyTorch-based variants.

Among all implementations studied in this paper, LSTM provided by Deeplea-rning4j emerges as the most optimal choice. It significantly enhances RePAD2's performance, achieving both high detection accuracy and efficient processing times across all tested time series datasets. Therefore, this combination is highly recommended for organizations and researchers seeking reliable and efficient anomaly detection solutions in real-time environments.

For our future work, we plan to release the source code of all implementations on a public software repository, such as GitHub, GitLab, or Bitbucket. We believe that our upcoming release will contribute to the advancement of the time series anomaly detection field by offering more effective and efficient approaches. Furthermore, we aim to further reduce RePAD2's false positives and deploy RePAD2 in various environments, such as Raspberry Pi to detect anomalies and intrusions on various IoT devices, mobile phones to identify anomalous activities or malicious behaviors, and Cyber-Physical Systems for more data-intensive and time-constrained anomaly and intrusion detection.

## Acknowledgement

The authors want to thank the anonymous reviewers for their reviews and valuable suggestions to this paper. This work has received funding from the Research Council of Norway through the SFI Norwegian Centre for Cybersecurity in Critical Sectors (NORCICS) project no. 310105.

## References

1. Abadi, M., Barham, P., Chen, J., Chen, Z., Davis, A., Dean, J., Devin, M., Ghemawat, S., Irving, G., Isard, M., et al.: Tensorflow: a system for large-scale machine learning. In: Osdi. vol. 16, pp. 265–283. Savannah, GA, USA (2016)
2. Ahmed, M., Mahmood, A.N., Hu, J.: A survey of network anomaly detection techniques. Journal of Network and Computer Applications **60**, 19–31 (2016)
3. Al-Ghuwairi, A.R., Sharrab, Y., Al-Fraihat, D., AlElaimat, M., Alsarhan, A., Algarni, A.: Intrusion detection in cloud computing based on time series anomalies utilizing machine learning. Journal of Cloud Computing **12**(1), 127 (2023)
4. Alimohammadi, H., Chen, S.N.: Performance evaluation of outlier detection techniques in production timeseries: A systematic review and meta-analysis. Expert Systems with Applications **191**, 116371 (2022)
5. Anton, S.D., Ahrens, L., Fraunholz, D., Schotten, H.D.: Time is of the essence: Machine learning-based intrusion detection in industrial time series data. In: 2018 IEEE International Conference on Data Mining Workshops (ICDMW). pp. 1–6. IEEE (2018)

6. Blázquez-García, A., Conde, A., Mori, U., Lozano, J.A.: A review on outlier/anomaly detection in time series data. ACM Computing Surveys (CSUR) **54**(3), 1–33 (2021)
7. Braei, M., Wagner, S.: Anomaly detection in univariate time-series: A survey on the state-of-the-art. arXiv preprint arXiv:2004.00433 (2020)
8. Cho, K., Van Merriënboer, B., Bahdanau, D., Bengio, Y.: On the properties of neural machine translation: Encoder-decoder approaches. arXiv preprint arXiv:1409.1259 (2014)
9. Chung, J., Gulcehre, C., Cho, K., Bengio, Y.: Empirical evaluation of gated recurrent neural networks on sequence modeling. arXiv preprint arXiv:1412.3555 (2014)
10. Deeplearning4j: Introduction to core Deeplearning4j concepts. https://deeplearning4j.konduit.ai/ (2023), [Online; accessed 231-July-2024]
11. Deka, P.K., Verma, Y., Bhutto, A.B., Elmroth, E., Bhuyan, M.: Semi-supervised range-based anomaly detection for cloud systems. IEEE Transactions on Network and Service Management (2022)
12. EarlyStopping: What is early stopping? https://deeplearning4j.konduit.ai/ (2023), [Online; accessed 31-July-2024]
13. Hochreiter, S., Schmidhuber, J.: Long short-term memory. Neural computation **9**(8), 1735–1780 (1997). https://doi.org/10.1162/neco.1997.9.8.1735
14. Hochreiter, S.: The vanishing gradient problem during learning recurrent neural nets and problem solutions. International Journal of Uncertainty, Fuzziness and Knowledge-Based Systems **6**(02), 107–116 (1998)
15. Hopfield, J.J.: Neural networks and physical systems with emergent collective computational abilities. Proceedings of the national academy of sciences **79**(8), 2554–2558 (1982)
16. Keras: Keras - a deep learning API written in python. https://keras.io/about/ (2023), [Online; accessed 31-July-2024]
17. Ketkar, N., Santana, E.: Deep learning with Python, vol. 1. Springer (2017)
18. Kieu, T., Yang, B., Jensen, C.S.: Outlier detection for multidimensional time series using deep neural networks. In: 2018 19th IEEE international conference on mobile data management (MDM). pp. 125–134. IEEE (2018)
19. Kovalev, V., Kalinovsky, A., Kovalev, S.: Deep learning with theano, torch, caffe, tensorflow, and deeplearning4j: Which one is the best in speed and accuracy? (2016)
20. Lee, M.C., Lin, J.C.: Impact of deep learning libraries on online adaptive lightweight time series anomaly detection. In: Proceedings of the 18th International Conference on Software Technologies - ICSOFT. pp. 106–116. INSTICC, SciTePress (2023). https://doi.org/10.5220/0012082900003538
21. Lee, M.C., Lin, J.C.: RePAD2: Real-time, lightweight, and adaptive anomaly detection for open-ended time series. In: Proceedings of the 8th International Conference on Internet of Things, Big Data and Security - IoTBDS. pp. 208–217. INSTICC, SciTePress. arXiv preprint arXiv:2303.00409. (2023)
22. Lee, M.C., Lin, J.C.: RoLA: A real-time online lightweight anomaly detection system for multivariate time series. In: Proceedings of the 18th International Conference on Software Technologies - ICSOFT. pp. 313–322. INSTICC, SciTePress (2023). https://doi.org/10.5220/0012077200003538
23. Lee, M.C., Lin, J.C., Gan, E.G.: ReRe: A lightweight real-time ready-to-go anomaly detection approach for time series. In: 2020 IEEE 44th Annual Computers, Software, and Applications Conference (COMPSAC). pp. 322–327. IEEE. arXiv preprint arXiv:2004.02319. (2020)

24. Lee, M.C., Lin, J.C., Gran, E.G.: RePAD: real-time proactive anomaly detection for time series. In: Advanced Information Networking and Applications: Proceedings of the 34th International Conference on Advanced Information Networking and Applications (AINA-2020). pp. 1291–1302. Springer. arXiv preprint arXiv:2001.08922. (2020)
25. Lee, M.C., Lin, J.C., Gran, E.G.: How far should we look back to achieve effective real-time time-series anomaly detection? In: Advanced Information Networking and Applications: Proceedings of the 35th International Conference on Advanced Information Networking and Applications (AINA-2021), Volume 1. pp. 136–148. Springer. arXiv preprint arXiv:2102.06560. (2021)
26. Lee, M.C., Lin, J.C., Gran, E.G.: SALAD: Self-adaptive lightweight anomaly detection for real-time recurrent time series. In: 2021 IEEE 45th Annual Computers, Software, and Applications Conference (COMPSAC). pp. 344–349. IEEE. arXiv preprint arXiv:2104.09968. (2021)
27. LinkedIn: linkedin/luminol [online code repository]. `https://github.com/linkedin/luminol` (2018), [Online; accessed 31-July-2024]
28. Moso, J.C., Cormier, S., de Runz, C., Fouchal, H., Wandeto, J.M.: Anomaly detection on data streams for smart agriculture. Agriculture **11**(11), 1083 (2021)
29. Nguyen, G., Dlugolinsky, S., Bobák, M., Tran, V., López García, Á., Heredia, I., Malík, P., Hluchỳ, L.: Machine learning and deep learning frameworks and libraries for large-scale data mining: a survey. Artificial Intelligence Review **52**, 77–124 (2019)
30. Paszke, A., Gross, S., Massa, F., Lerer, A., Bradbury, J., Chanan, G., Killeen, T., Lin, Z., Gimelshein, N., Antiga, L., et al.: Pytorch: An imperative style, high-performance deep learning library. Advances in neural information processing systems **32** (2019)
31. Pereira, J., Silveira, M.: Learning representations from healthcare time series data for unsupervised anomaly detection. In: 2019 IEEE international conference on big data and smart computing (BigComp). pp. 1–7. IEEE (2019)
32. Ren, H., Xu, B., Wang, Y., Yi, C., Huang, C., Kou, X., Xing, T., Yang, M., Tong, J., Zhang, Q.: Time-series anomaly detection service at microsoft. In: Proceedings of the 25th ACM SIGKDD international conference on knowledge discovery & data mining. pp. 3009–3017 (2019)
33. Schneider, J., Wenig, P., Papenbrock, T.: Distributed detection of sequential anomalies in univariate time series. The VLDB Journal **30**(4), 579–602 (2021)
34. Twitter: AnomalyDetection R package [online code repository]. `https://github.com/twitter/AnomalyDetection` (2015), [Online; accessed 31-July-2024]
35. Vito, S.: Air Quality. UCI Machine Learning Repository (2016), DOI: `10.24432/C59K5F`
36. Wang, Z., Liu, K., Li, J., Zhu, Y., Zhang, Y.: Various frameworks and libraries of machine learning and deep learning: a survey. Archives of computational methods in engineering pp. 1–24 (2019)
37. Yatish, H., Swamy, S.: Recent trends in time series forecasting–a survey. International Research Journal of Engineering and Technology (IRJET) **7**(04), 5623–5628 (2020)
38. Zahidi, Y., El Younoussi, Y., Al-Amrani, Y.: A powerful comparison of deep learning frameworks for arabic sentiment analysis. International Journal of Electrical & Computer Engineering (2088-8708) **11**(1) (2021)
39. Zhang, J.E., Wu, D., Boulet, B.: Time series anomaly detection for smart grids: A survey. In: 2021 IEEE Electrical Power and Energy Conference (EPEC). pp. 125–130. IEEE (2021)

40. Zhang, Q., Li, X., Che, X., Ma, X., Zhou, A., Xu, M., Wang, S., Ma, Y., Liu, X.: A comprehensive benchmark of deep learning libraries on mobile devices. In: Proceedings of the ACM Web Conference 2022. pp. 3298–3307 (2022)
41. Zhang, X., Wang, Y., Shi, W.: pcamp: Performance comparison of machine learning packages on the edges. In: HotEdge (2018)